\documentclass{article}
\usepackage{graphicx} 
\usepackage[backend=biber,maxbibnames=19]{biblatex}
\usepackage{hyperref}
\usepackage{xurl}
\hypersetup{breaklinks=true}
\title{ZeroShotDataAug: Generating and Augmenting Training Data with ChatGPT}

\author{
Solomon Ubani
\and
Suleyman Olcay Polat
\and
Rodney D Nielsen \and {\{solomon.ubani, suleymanolcay.polat, rodney.nielsen\}}@unt.edu
}

\begin{document}

\maketitle

\begin{abstract}
In this paper, we investigate the use of data obtained from prompting a large generative language model, ChatGPT, to generate synthetic training data with the aim of augmenting data in low resource scenarios. We show that with appropriate task-specific ChatGPT prompts, we outperform the most popular existing approaches for such data augmentation. Furthermore, we investigate methodologies for evaluating the similarity of the augmented data generated from ChatGPT with the aim of validating and assessing the quality of the data generated. 
\end{abstract} 

\section{Introduction}

Data augmentation is a technique to increase the size of the training data available to machine learning models without requiring additional human annotation of data. Increasing the size of training data, provided the additional data is somewhat diverse, is pertinent to enable model generalization especially in low resource tasks. The aim of this paper is to evaluate zero-shot prompting of ChatGPT for data augmentation in the low resource scenario.
\\ \\
Wei and Zou \cite{eda} proposed Easy Data Augmentation (EDA) which is a technique based on word replacement that includes four types of operations: synonym replacement, random insertion, random deletion, and random swap. In synonym replacement, words with similar meanings are substituted for some of the original words in the text. This helps to introduce variations in the text and expand the range of vocabulary. Random insertion involves adding new words to the text, which are not present in the original data. This helps to increase the diversity of the text and can also help models learn to deal with out-of-vocabulary words. In random deletion, words are randomly removed from the text. This can help to simulate situations where some words may be missing in the input and can help the model become more robust to noise. In random swap, two words in the text are randomly swapped. This can help to introduce variations in the text and improve the diversity of the training data.
\\ \\ 
Back translation \cite{backtranslation} is a common data augmentation technique in Natural Language Processing (NLP). It involves translating a sentence or text from one language to another and then translating it back to the original language using a machine translation model. Back translation can introduce variations in the text which can help to create more diverse and representative data for training NLP models. Researchers have also used pretrained  autoencoder transformer models like BERT \cite{bert}, CBERT \cite{cbert}, and BART \cite{bart} to augment text data in NLP \cite{kumarpaper} \cite{cbert}. These techniques generally mask some words in the training set and utilize the pretrained models to predict the masked word(s). This could create more diverse data since the predicted words could vary from the original word. The authors include the class labels during finetuning and language modelling to aid the models in predicting the masked words in the context of their labels. Kumar et al. \cite{kumarpaper} also utilzed an autoregressive pretrained language model, GPT-2 \cite{gpt2}, to augment data by prompting GPT-2 to complete the sentence given only the first few words of the sentence and the training data label.
\\ \\
Dai et al. \cite{fewshotchatgpt} utilized few-shot prompting of ChatGPT \cite{chatgpt} for data augmentation to produce several variations of each sentence in the training sample. The generated sentences are similar in meaning but have different syntax. ChatGPT is a conversational agent that utilizes OpenAI's GPT-3.5 \cite{gpt35}---a large-scale language model trained on a vast corpus of diverse and information-rich web data and Reinforcement Learning from Human Feedback (RLHF). Few-shot prompting is a technique that enables a language model to perform a new task with only a few examples of training data. Zero-shot prompting is a technique in which a language model is provided with a task description, rather than direct supervision or training data, to perform a specific task. The task description is in the form of a prompt or a question that guides the model on how to generate the desired output. In this paper, we investigate zero-shot prompting of ChatGPT for data augmentation. \\ \\

The main contributions of this paper are: 

\begin{itemize}
  \item Evaluation of zero-shot prompting of ChatGPT for data augmentation on three datasets. 

  \item  Three methodologies to evaluate the similarity of the data generated from zero-shot prompting of ChatGPT with the training and test sets with the aim of validating and assessing the quality of the data generated.

  \item  Investigation of the marginal returns of data generated from different data augmentation techniques. 
\end{itemize}

\section{Datasets}

The datasets we use to evaluate our data augmentation methodology are popular benchmark natural language understanding datasets that researchers have utilized to evaluate other data augmentation techniques. We evaluate on three text classification datasets:

\begin{itemize}

\item \textbf{SST-2 \cite{sst2}}: Stanford Sentiment Treebank consists of movie reviews from the Rotten Tomatoes website, where each example in the reviews is labeled with its sentiment polarity (positive or negative).

\item \textbf{SNIPS \cite{snips}}: The Spoken Natural Language Interaction for Personal Services dataset consists of annotated spoken queries related to seven intents in the domains of music, weather, and home automation.

\item \textbf{TREC \cite{trec}}: A question classification dataset sourced from the Text Retrieval Conference. It contains six question types (whether the question is about an abbreviation, description, entity, human, location, or numeric value).

 
\end{itemize}
\subsection{Low-Resource Data Scenario}

In this research work, we evaluate the impact of data augmentation in the low-data scenario. We follow a similar approach to previous work \cite{kumarpaper} on data augmentation by randomly subsampling only 10 examples per class on each task for both the training and the development sets. 
\\ \\ 
To improve the model’s performance on the low training data scenario, for each task, we incorporate the synthetic data generated by either ChatGPT or the comparison approaches. Subsequently, we assess the models' performance on the entire test set. To address any stochastic variation, we repeat all experiments 15 times. 

\section{Methods}

\subsection{Baseline Methods}

In this research, we compare zero-shot prompting of ChatGPT with
other popular data augmentation methods listed below (see original sources cited following the method names for further details): 

\begin{itemize}
\item \textbf{EDA} \cite{eda}: We used the random word replacement techniques that include four types of operations: synonym replacement, random insertion, random deletion, and random swap. We set $\alpha$ (proportion of words replaced) to 0.10 since the examples are short sentences.

\item \textbf{BackTrans} \cite{backtranslation}: We translated the training example from one language to another and then translated it back to the original language using a machine translation model.\footnote{Google Translate (https://pypi.org/project/googletrans/).} 

\item \textbf{CBERT} \cite{cbert}: First, we utilize BERT’s segment embeddings to condition the BERT model on the class labels during finetuning.\footnote{In all baseline experiments that require finetuning, we trained for 20 epochs using AdamW optimizer with a learning rate of $5\times10^{-4}$ and the cross entropy loss.} We then finetuned the model with the masked language model (MLM) objective which randomly masks some words in the sequences and aims to predict the original word using the context. Finally, we use the resulting model to predict and replace masked words in the training set. 

\item \textbf{BERTexpand} \cite{kumarpaper}: First, we prepended the label to each sequence in the training data and added the labels to the model vocabulary before finetuning. We then finetuned the model with the MLM objective. Finally, we use the resulting model to predict and replace masked words in the training set.

\item \textbf{BERTprepend} \cite{kumarpaper}: First, we prepended the label to each sequence in the training data without adding the labels to the model vocabulary before finetuning. We then finetuned the model with the MLM objective. Finally, we used the resulting model to predict and replace masked words in the training set.

\item \textbf{GPT2context} \cite{kumarpaper}: First, we prepended the label to each sequence in the training data before finetuning GPT-2. Next, We then finetuned the GPT-2 model on the MLM objective. Finally, we prompted the resulting model to complete the sentences given only the first three words of the sentence prepended by the label in the training set. 

\item \textbf{BARTword} \cite{kumarpaper}: First, we prepended the label to each sequence in the training data without adding the labels to the model vocabulary. Next, we finetuned the BART model on the denoising reconstruction task where 40\% of words are masked and the goal of the model is to reconstruct the original sequence. Finally, we used the resulting model to predict and replace the masked word in each sequence of the training set.

\item \textbf{BARTspan} \cite{kumarpaper}: First, we prepended the label to each sequence in the training data without adding the labels to the model vocabulary. Next, we finetuned the BART model on the denoising reconstruction task where 40\% of words are masked and the goal of the model is to reconstruct the original sequence. Finally, we used the resulting model to predict and replace the masked spans of words in the training set.

\item \textbf{ChatGPTfew-shot} \cite{fewshotchatgpt}: We used few-shot prompting of ChatGPT for data augmentation to produce several paraphrases of each sentence in the training set.

\end{itemize}

\subsection{Prompts for Zero-shot Data Augmentation}

In this section, we list the prompts used to generate augmented examples for each class. The prompts were generated by observing the task, class description and five instances per class of the training data. 
\\ \\
\textbf{Dataset: SST-2}
\begin{itemize}

\item  []
\textbf{Class}: Positive \\
\textbf{Prompt}: Generate 20 sentences that are positive reviews to a movie
\textbf{Class}: Negative \\
\textbf{Prompt}: Generate 20 sentences that are negative reviews to a movie
\end{itemize}
\textbf{Dataset: SNIPS}

\begin{itemize}
\item[]  \textbf{Class}: RateBook \\ 
{\textbf{Prompt}: Generate 20 sentences in an imperative mood where a human tells a digital assistant to rate a random book and the human provides the numerical rating. Use random book names. Do not mention the name of the digital assistant.}
\\ 
\textbf{Class}: AddToPlaylist \\
\textbf{Prompt}: Generate 20 sentences in an imperative mood where a human tells a digital assistant to add music to a playlist and the human provides the music name. Use random music and playlist names. Do not mention the name of the digital assistant. 
\\ 
\textbf{Class}: PlayMusic \\
\textbf{Prompt}: Generate 20 sentences in an imperative mood where a human tells a digital assistant to play a music and the human provides the music name. Use random music and names. Do not mention the name of the digital assistant.
\\ 
\textbf{Class}: BookRestaurant \\
\textbf{Prompt}: Generate 20 sentences in an imperative mood where a human tells a digital assistant to book a restaurant and the human provides the restaurant or food name. Use random restaurant and food names. Do not mention the name of the digital assistant.
\\ 
\textbf{Class}: GetWeather \\
\textbf{Prompt}: Generate 20 sentences in an imperative mood where a human asks a digital assistant about the weather. Sometimes the human may provide the time and city. Use random city names. Do not mention the name of the digital assistant.
\\ 
\textbf{Class}: SearchCreativeWork \\
\textbf{Prompt}: Generate 20 sentences in an imperative mood where a human asks a digital assistant to find a specific creative work. The creative work could be a movie, tv show, book or game. Use random movie names, tv shows names, books, games. Sometimes the human asks for a specific creative work. Do not mention the name of the digital assistant.
\\ 
\textbf{Class}: SearchScreeningEvent \\
\textbf{Prompt}: Generate 20 sentences in an imperative mood where a human asks a digital assistant to find information about a movie or screening in the theater. Sometimes the human asks for a specific movie. Do not mention the name of the digital assistant.
\end{itemize}

\textbf{Dataset: TREC}
\begin{itemize}

\item[] \textbf{Class}: Abbreviation \\
\textbf{Prompt}: Generate 20 questions asking about the meaning of an abbreviation
\\ 
\textbf{Class}: Entity \\
\textbf{Prompt}: Generate 20 questions asking about a random example of a noun or entity. Actually use different nouns or entities in each sentence. 
\\ 
\textbf{Class}: Description \\
\textbf{Prompt}: Generate 20 sentences that are only "what is" questions that query for a definition.
\\
\textbf{Class}: Human \\
\textbf{Prompt}: Generate 20 questions about random facts about a person or people in history.
\\
\textbf{Class}: Location \\
\textbf{Prompt}: Generate 20 sentences that are questions that ask the location of a place in history. Use a different place for each sentence
\\
\textbf{Class}: Numeric Value \\
\textbf{Prompt}: Generate 20 sentences that are questions about a numeric fact in history
\end{itemize}
\subsection{Evaluating the Similarity of the Generated Data from ChatGPT Versus the Training and Test Data}
Since there is a chance that ChatGPT might have been trained on the datasets for some of our tasks, we investigate data contamination in all our datasets. We investigate data contamination by measuring the similarity between the data generated from ChatGPT and the datasets for each of our tasks using three similarity metrics: Cosine(Sentence Embedding), Cosine(TF-IDF), and Percent Word Overlap (each detail later in this section). 

First, for each task, we compared each example generated by ChatGPT with all the examples from the training set and, separately, with the testing set. For each generated example, we selected the maximum similarity score and then averaged the scores over the generated dataset. 
\\ \\
The three metrics used to evaluate similarity were:
\\
\begin{itemize}
  \item \textbf{Sentence Embedding}: 
  We used the MiniLM model \cite{minilm} in the sentence transformer library \cite{sentencetransformers} to obtain the embedding of each example. We calculated the cosine similarity between the embeddings of the pair of examples.
  \item \textbf{TF-IDF}:
  We used TF-IDF vectors to calculate the similarity between examples. First, we removed the stop words from each example and created a corpus by combining the training data, testing data and zero-shot ChatGPT generated data. We used the resulting corpus as the corpus for the TF-IDF vector similarity approach. Finally, we converted each example into a TF-IDF vector and calculated the cosine similarity between each example.
  \item \textbf{Word Overlap}:
  We calculated the word overlap scores between examples to obtain their similarity. First, we removed the stop words and punctuation from all the examples. Next, for a example pair, we counted the number of unique words that appear in both examples and divided it by the number of unique words that appear on the longer example of the pair. This metric gives us the percentage of overlapping words between the two examples.
\end{itemize}

\begin{table}[htbp]
  \centering
    \begin{tabular}{|p{15.68em}|c|c|c|}
    \hline
  \multicolumn{1}{|r|}{} & \multicolumn{1}{p{6.135em}|}{\textbf{Sentence Embedding}} & \multicolumn{1}{p{4.035em}|}{\textbf{TF-IDF}} & \multicolumn{1}{p{4.38em}|}{\textbf{Word Overlap}} \\
    \hline
    \hline
    ChatGPTzero-shot to SNIPStest & 0.553 & 0.239 & 0.265 \\
    \hline
    SNIPStrain to SNIPStest & 0.593 & 0.362 & 0.426 \\
    \hline
    \hline
    ChatGPTzero-shot to TRECtest & 0.528 & 0.330 & 0.271 \\
    \hline
    TRECtrain to TRECtest & 0.448 & 0.240 & 0.202 \\
    \hline
    \hline
    ChatGPTzero-shot to SST-2test & 0.600 & 0.293 & 0.271 \\
    \hline
    SST-2train to SST-2test & 0.535 & 0.229 & 0.211 \\
    \hline
    \end{tabular}%
  \caption{Data augmentation similarity results of ChatGPTzero-shot versus the original training sets relative to the testing sets.}
\label{tab:similaritytable}%
\end{table}%

\begin{table}[htbp]
  \centering
    \begin{tabular}{|p{15.68em}|c|c|c|}
    \hline
  \multicolumn{1}{|r|}{} & \multicolumn{1}{p{6.135em}|}{\textbf{Sentence Embedding}} & \multicolumn{1}{p{4.035em}|}{\textbf{TF-IDF}} & \multicolumn{1}{p{4.38em}|}{\textbf{Word Overlap}} \\
    \hline
    ChatGPTzero-shot to SNIPStrain & 0.629 & 0.341 & 0.360 \\
    \hline
    ChatGPTzero-shot to TRECtrain & 0.634 & 0.435  & 0.404 \\
    \hline
    ChatGPTzero-shot to SST-2train & 0.635 & 0.351   & 0.317 \\
    \hline
    \end{tabular}%
  \caption{Data augmentation similarity results between ChatGPTzero-shot and training sets.}
\label{tab:similaritytable2}%
\end{table}%

\begin{table}[htbp]
  \centering
    \begin{tabular}{|l|c|c|c|}
    \hline
          & \multicolumn{3}{c|}{\textbf{Word Overlap Similarity}} \\
    \cline{2-4}
          & $>$ 66\% & Max \% & \# Exs at Max \\ 
    \hline
    \hline
    ChatGPTzero-shot to SNIPStest & ~0.7     & ~67     & 2 \\
    \hline
    SNIPStrain to SNIPStest & 11.0     & 100     & 142 \\
    \hline
    \hline
    ChatGPTzero-shot to TRECtest & ~5.0     & 100     & 3 \\
   \hline
    TRECtrain to TRECtest & ~1.3     & 100     & 14 \\
    \hline
    \hline
    ChatGPTzero-shot to SST-2test & ~0.0     & ~43     & 1 \\
    \hline
    SST-2train to SST-2test  & ~0.2     & 100     & 4 \\
    \hline
    \end{tabular}%
        \caption{Three statistics for the word overlap similarity: the percentage of examples where word overlap similarity is greater than 66\%, the maximum word overlap similarity, and the number of examples with the maximum overlap}
\label{tab:wordoverlaptable}%
\end{table}%

As seen in Tables \ref{tab:similaritytable}, \ref{tab:similaritytable2} and \ref{tab:wordoverlaptable} the results from the data similarity investigation revealed that there is little to no data contamination from the data generated by ChatGPT for any of our tasks. The investigation also reveals that the data generated by ChatGPT is on average 7\% more similar to the training datasets than the testing datasets. We believe this difference stems from the higher number of examples in the training datasets relative to testing. Hence, since we are calculating the maximum similarity between each example and the entire dataset, a larger dataset increases the probability that there will be an example with a high similarity.

As seen in Table \ref{tab:wordoverlaptable}, for SNIPS percent word overlap, the original training data is 16\% more similar to the testing data than is the generated data. 
The most similar generated examples to the testing set matched two out of three content words (two examples); whereas, the most similar original training examples had a 100\% content word overlap with a testing example (142 examples).
Together, these statistics imply the generation did not rely on memorization by the underlying pretrained LLM (ChatGPT) for this task.

For the TREC and SST-2 testing data, the generated training data appeared more similar than the original training data by 7\% and 6\%, respectively. 
For TREC, there were 3 generated examples and 14 original training examples having complete content word overlap with testing examples. For the generated examples, this complete overlap is associated with common very short, questions (``Who invented the telephone?", ``What year did the Titanic sink?", ``What is a tsunami?"). Given the small number of such questions, we don’t see this as a worrisome problem.
For SST-2, the maximum overlap with a testing example for the generated data was only 43\% (in one example); whereas, there were four original training examples with 100\% content word overlap with a testing example.

Again, these statistics imply ChatGPT did not rely on memorization of the testing data during its pretraining. Hence, results reported in the following sections should be predictive of what others will achieve on new tasks that are relatively similar in nature.

\subsection{Model Implementation}
We finetuned the pretrained BERT-base uncased model for all three text classification tasks. The hyperparameters we used to finetune BERT were similar to the hyperparameters used by Kumar et al. \cite{kumarpaper} for a fair comparison. The hyperparameters were a training time of twelve epochs, batch size of 32, the AdamW optimizer function with a learning rate of $5\times10^{-5}$, and the categorical cross entropy loss function. During the training the models are saved after each epoch and the best performing model on the development set is chosen to be run on the testing set for each model.

\section{Results} 

As seen in Table \ref{tab:resultstable}, with zero-shot prompting of ChatGPT for data augmentation, our model achieves an accuracy of 78.1\%, 91.2\%, and 75.2\% for SST-2, SNIPS, and TREC respectively. Zero-shot prompting of ChatGPT outperformed all existing data augmentation methods on SST-2 and TREC while showing similar performance to few-shot prompting of ChatGPT on SNIPS. Specifically, our model surpassed the best non-ChatGPT-based model by 20\%, 4\%, and 8\% on SST-2, SNIPS, and TREC, respectively. It is also noteworthy that ChatGPTzero-shot had the lowest variance for both SNIPS and TREC, and the second lowest for SST-2.

\begin{table}[htbp]
  \centering
    \begin{tabular}{|l|c|c|c|}
    \hline
    \textbf{Model} & \textbf{SST-2} & \textbf{SNIPS} & \textbf{TREC} \\
    \hline
    No Aug & 52.9 (5.0) & 79.4 (3.2) & 48.6 (11.5) \\
    \hline
    EDA   & 53.8 (4.4) & 85.8 (3.0) & 52.6 (10.5) \\
    \hline
    BackTrans. & 57.5 (5.6) & 86.5 (2.4) & 66.2 (~8.5) \\
    \hline
    CBERT & 57.4 (6.7) & 85.8 (3.5) & 64.3 (10.9) \\
    \hline
    BERTexpand & 56.3 (6.5) & 86.1 (2.7) & 65.3 (~6.1) \\
    \hline
    BERTprepend & 56.1 (6.3) & 86.8 (1.6) & 64.7 (~9.6) \\
    \hline
    GPT2context & 55.4 (6.7) & 86.6 (2.7) & 54.3 (10.1) \\
    \hline
    BARTword & 58.0 (6.8) & 86.8 (2.6) & 63.7 (~9.8) \\
    \hline
    BARTspan & 57.7 (7.1) & 87.2 (1.4) & 67.3 (~6.1) \\
    \hline
    ChatGPTfew-shot & 69.6 (5.8) & \textbf{91.3} (1.4) & 66.7 (~8.0) \\
    \hline
    ChatGPTzero-shot &\textbf{78.1} (5.1) & 91.2 (1.3) & \textbf{75.3} (~4.0) \\
    \hline
    \end{tabular}%
  \label{tab:resultstable}%
  \caption{Accuracy (and standard deviation) for each data augmentation method. All results except for ChatGPTfew-shot and ChatGPTzero-shot were the same results reported by Kumar et al.~\cite{kumarpaper}.}
\end{table}%

\section{Evaluating the Performance Improvement with more Augmentation Examples}

In this section, we investigate how performance is impacted by the number of examples created through augmentation for all data augmentation methods. In Figure 1, we show the performance of different data augmentation techniques on the SST-2 dataset for $K$ augmentation examples per original training example.\footnote{For Back Translation, we used $K$ different languages.} 

\begin{figure}[htp]
 \includegraphics[width=1\textwidth]{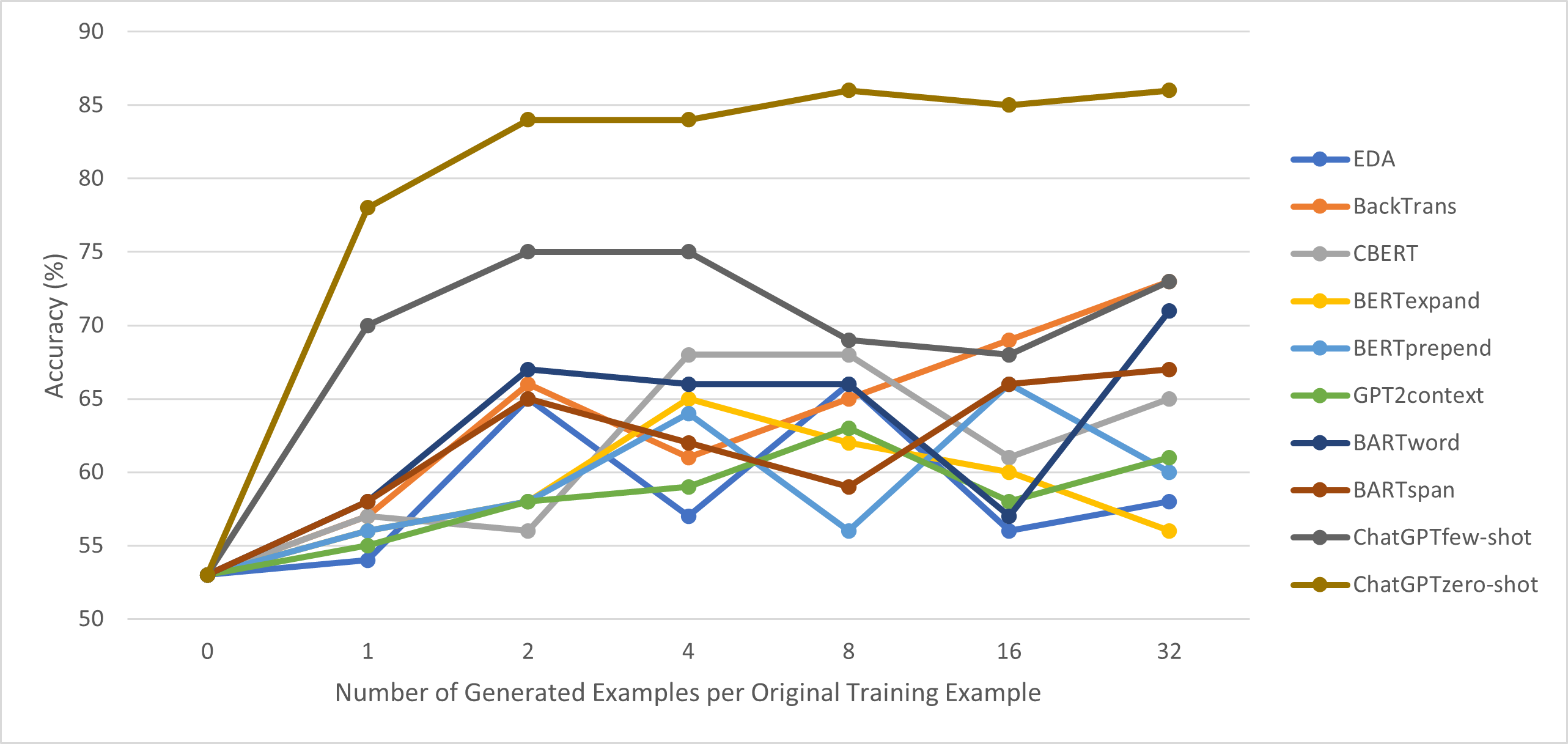}
    \caption{Accuracy as the number of generated examples per original training example increases}
    \label{fig:9.1}
\end{figure}

As seen in Figure \ref{fig:9.1}, ChatGPTzero-shot outperforms all other data augmentation methods for $K \in \{1, 2, 4, 8, 16, 32\}$. This provides further evidence on the effectiveness of using ChatGPT for data augmentation in low resource scenarios. 

\section{No Training Data Scenario}

In this section, we investigate the performance of our model on SST-2, SNIPS and TREC's test dataset when we use augmented data from zero-shot prompting of ChatGPT without any data from the original training datasets. We use the same prompts described in section 3.2 to generate 20 instances per class. We train the models using the same hyperparameters as in section 3.4. We repeat both steps 15 times to account for stochasticity. Our model obtained a mean accuracy of 0.80, 0.78, and 0.62 on SST-2, SNIPS and TREC's designated test sets respectively. Interestingly, with no data from the original training set, our model outperforms all existing augmentation methods on SST-2 and within a standard deviation of the best result on TREC. These results further highlight the significance of zero-shot prompting of ChatGPT. Unlike other data augmentation techniques, our method of data augmentation can be used even in the absence of any training data. 

\section{Discussion} 
This research provides an easy-to-use and intuitive methodology for generating augmented data. In our experiments, our method of augmenting training data for these NLP tasks by zero-shot prompting ChatGPT shows great promise in the low resource scenario, substantially outperforming all but one of the baseline methods -- few-shot ChatGPT, which only exceeded it by 0.1\% on one task. It should however be noted that, as with other data augmentation methods, zero-shot prompting of ChatGPT for data augmentation does not necessarily improve results where there is a large training dataset. 
\\ \\
One of the limitations of previous data augmentation methods is that the quality of augmented data strongly depends on the original training dataset. The quality of data generated from zero-shot prompting of ChatGPT is not limited by the human-annotated training data. This research also provides evidence that data generated from zero-shot prompting of ChatGPT has slower diminishing returns compared to many existing data augmentation techniques. Furthermore, on SST-2, our model achieved better results using only data from zero-shot prompting of ChatGPT than did other data augmentation approaches that supplemented an initial human-annotated training dataset. This further highlights the effectiveness of zero-shot prompting of ChatGPT as a data augmentation approach. 
\\ \\ 
It is important to note that the quality of augmented data generated by zero-shot prompting of ChatGPT depends on the quality of the prompts. The prompts in this research were human generated based on the task description and observing a few training data instances. While there is a lot of current research in the area of prompt engineering, there are still no task-independent well-established best practices for how to generate effective prompts. This research presents a methodology for evaluating the augmented data generated from large language models. To evaluate the augmented data, we calculated the sentence embedding similarity, TF-IDF vector similarity, and word overlap scores of the synthetic examples compared to all the examples in the training and test data. This revealed that there was very little data generated with high similarity scores, making it unlikely this stemmed from memorization of the datasets by ChatGPT during its training.

\section{Conclusion and Future Work}

This paper proposes a novel technique for generating augmented data for machine learning tasks using zero-shot prompting of ChatGPT. The experimental results demonstrate that the proposed method substantially outperforms all the baseline approaches for the three tasks investigated in this research, with the single exception of few-shot ChatGPT performing as well on one task. This indicates our method's potential as a promising data augmentation method in the low resource setting. 

The data augmentation approach investigated in this research relies on manually engineering effective prompts for each task which requires some expertise. Future researchers can explore more systematic approaches to prompt engineering especially for tasks that cannot be adequately described within a concise prompt of one to three sentences.


\printbibliography

\end{document}